\documentclass{article}

\usepackage{amsmath}
\usepackage{multicol,lipsum}
\usepackage{enumerate}
\usepackage{graphicx}
\usepackage{amssymb}
\usepackage{comment}
\usepackage{caption}
\usepackage[lined,algonl,boxed]{algorithm2e}
\usepackage{subcaption}
\usepackage[table]{xcolor}
\usepackage{slashbox}
\usepackage{style}

\title{Markov Switching Model for Driver Behavior Prediction: Use cases on Smartphones}

\author{
  Ahmed~B.~Zaky\\
  Faculty of Engineering (Shoubra), Benha University, Cairo 11240, Egypt\\
  Big Data Institute, College of Computer Science and Software Engineering\\
  Shenzhen University, Shenzhen 518060, Guangdong, China \\
  \texttt{ahmed.zaky@feng.bu.edu.eg} \\
  \texttt{ahmed.zaky@szu.edu} \\
  \And
  Mohamed A. Khamis \\
  Cyber-Physical Systems Lab\\
  Egypt-Japan University of Science and Technology (E-JUST)\\
  New Borg El-Arab City, Postal Code 21934, Alexandria, Egypt\\
  \texttt{mohamed.khamis@ejust.edu.eg} \\
  \AND
  Walid Gomaa \\
  Cyber-Physical Systems Lab\\
  Egypt-Japan University of Science and Technology (E-JUST)\\
  New Borg El-Arab City, Postal Code 21934, Alexandria, Egypt\\
  Faculty of Engineering, Alexandria University, Alexandria 21544, Egypt\\
  \texttt{walid.gomaa@ejust.edu.eg} \\
}

\begin{document}








\maketitle

\begin{abstract}



Several intelligent transportation systems focus on studying the various driver behaviors for numerous objectives.
This includes the ability to analyze driver actions, sensitivity, distraction, and response time. 
As the data collection is one of the major concerns for learning and validating different driving situations,
we present a driver behavior switching model validated by a low-cost data collection solution using smartphones.
The proposed model is validated using a real dataset to predict the driver behavior in short duration periods. 
A literature survey on motion detection (specifically driving behavior detection using smartphones) is presented.
Multiple Markov Switching Variable Auto-Regression (MSVAR) models are implemented to achieve a sophisticated fitting with the collected driver behavior data. 
This yields more accurate predictions not only for driver behavior but also for the entire driving situation.
The performance of the presented models together with a suitable model selection criteria is also presented.
The proposed driver behavior prediction framework can potentially be used in accident prediction and driver safety systems.
\end{abstract}



\section{Introduction}
\label{sec:introduction}


Models for collision avoidance systems, congestion assistants, crash prediction software, etc. have been developed to support a vehicle driver while doing complicated tasks. Accordingly, these systems increase the safety limit for both the drivers and pedestrians.
Fully automated autonomous driving systems still need more attention in building models for different driving assistance tasks.
Developing a driving behavior model that can be adapted to different driving situations and be able to cover most driving behaviors, is still a challenging task.

Machine learning (ML) is one of the fastest growing areas of science.
It has been used in many applications;
e.g., traffic signal control \cite{Khamis2014AdaptiveMultiObjectiveReinforcementLearning,Khamis2012AdaptiveTrafficControl}, classification, and recognition.
ML techniques such as regression models \cite{angkititrakul2013stochastic}, neural networks (NNs) \cite{panwai2007neural}, and fuzzy systems \cite{ma2006neural} have been used recently in modeling patterns of driving situations.
However, such models face the complication of understanding different driving situations (especially the unexpected ones).
Driving tasks can be segmented into driving regimes mapped to different driving situations with different response for each driver.
A driver usually switches between different behaviors such as car following, lane changing, mobile messaging, sign reading, etc.
It is normal to see a driver perform more than one task at the same time, e.g., following a car while switching radio channels or messaging, etc.
In this article, we propose a stochastic model that is suitable for detecting and classifying different driving regimes.

Preliminary results of the work presented in this paper have been published in \cite{Zaky2014CarFollowingRegimeTaxonomyMarkovSwitching, Zaky2015CarFollowingMarkovRegimeClassification}.
In this paper, the Expectation Maximization (EM) and Markov Chain Monte Carlo (MCMC) are used for estimating the proposed model parameters.
Moreover, we calibrate the model for car following driver behavior using our own data set collected by smartphones (as a low cost solution for collecting driving data) plus a naturalistic driving data set presented in \cite{manstetten1997traffic}.
We also present a brief survey for different machine learning models employed for driver behavior and data collection based on smartphones.

The rest part of this paper is organized as follows.
Section \ref{sec:RelatedWork} presents the state of the art literature review.
Specifically, we focus on motion detection with use cases on driving behavior detection using smartphones. 
Also, this section provides the necessary background for the work presented in this paper. This includes detailed description for driver behavior models
Section \ref{sec:Methodology} introduces Markov Switching Vector Auto-Regressive model and Bayesian Gipps sampling for model parameters estimation.
Section \ref{sec:Experimentation} depicts the data collection process using smartphones, the car following dataset, and  the adopted driving behavior model.
Section \ref{sec:ResultsAndDiscussion} presents the results of using both of the data collected using smartphones and the naturalistic driving data.
Finally, Section \ref{sec:ConclusionsAndFutureWork} concludes the work presented in this article and provides directions for future research.

\section{Related Work}
\label{sec:RelatedWork}
\subsection{Driver behavior models}
Recently, machine learning approaches have been proposed for driver behavior modeling.
Car following is the most popular behavior for evaluating these approaches.
Three models are mainly used:
Hidden Markov Models (HMM), Gaussian Mixture Models (GMM), and Piece-Wise Auto Regressive Exogenous models (PWARX).
These models have achieved remarkable results in simulating driving scenarios.
Additionally, these models divide each complex driving pattern into sub-patterns using mixture components.
Models introduce different methods for calculating the latent classes, the relationship between the observed variables and each class, the estimation of class parameters, and the number of latent classes.

\subsubsection{Hidden Markov models}
HMM has been used for driver behavior modeling in different situations such as the model implemented in \cite{dapzol2005driver}.
This model uses sensor data evolution to predict the real current driving situation.
The results achieved a prediction accuracy of 80\% of driver behavior recognition from the initial driver movements.
In \cite{2011TISCI}, the authors presented a collision warning system based on HMM.
Traffic models based on HMM have been reviewed in \cite{dapzol2005driver} and \cite{sathyanarayana2008driver}.

\subsubsection{Gaussian mixture models}
A stochastic driver behavior modeling framework based on GMM is presented in \cite{angkititrakul2013stochastic}. 
The model calculates the joint probability distribution for a number of driving signals (following distance, vehicle velocity, brake and gas pedal forces, and vehicle dynamics).
The model implements two GMMs as a representation of gas and brake pedals, and their relation with the follower velocity and the gap distance. 
A main issue with the GMM is the selection of the model components, the authors proposed the use of a Dirichlet process as a non-parametric Bayesian approach that selects the optimal number of model components.
The model fits the driving observations for each driver. In addition, a general driver model is implemented based on fitting the observations of several drivers. 
The authors used different mixtures (4, 8, 16, and 32) for evaluating the model performance.

\subsubsection{Piece-wise auto-regressive exogenous models (PWARX)}
PWARX have been presented in \cite{sekizawa2007modeling} and \cite{okuda2009multi} to model human driving behavior as a Hybrid Dynamical System (HDS). 
The proposed approach is switching between simple linear behavior models instead of modeling a non-linear complex model.
The driver behavior recognition model introduced in \cite{sekizawa2007modeling} is a standard HMM extended by embedding an auto-regressive exogenous model (ARX) in each discrete state. The authors introduced a simulation of a collision avoidance system.

In \cite{akita2008analysis}, a car following model classification approach has been introduced based on PWARX as a segmentation approach and a $K$-means clustering for the input vector. The classification between modes is done using a Support Vector Machine (SVM).
The PWARX models have impressive results in modeling driver behavior.
However, the PWARX models have two problems \cite{takeda2016modeling}.
First, the model can not classify and estimate the behavior simultaneously.
Second, it is unable to handle a probabilistic time varying data.
The Probability weighted Auto-Regressive model (PrARX) proposed in \cite{okuda2013modeling} is an extension to PWARX. 
The PrARX model addresses these two issues by composing multiple ARX models by a probabilistic weighting function.

The PWARX and GMM models are compared in \cite{takeda2016modeling} for car following behavior modelling.
These models predict a pattern of brake and gas pedal response of the driver as a feedback to the current vehicle velocity and gap distance between vehicles.
The results showed that the PWARX approach outperforms GMM in all cases.
The prediction of gas pedal behavior was better than the prediction of brake pedal in both models.
The results also showed that the operation of gas pedal was smoother than that of the brake pedal.
In addition, the gas pedal took longer control time than the pulse-wise shorter period brake pedal.

\subsection{Driving behavior detection using smartphones}
Motion detection in traffic networks (e.g., anomaly detection, car following driver behavior, etc.) has gained much concern in the last few years.
In \cite{Zaky2014CarFollowingRegimeTaxonomyMarkovSwitching}, the authors proposed a Markov regime switching-based model to estimate the driver behavior and extract different driving regimes.
The proposed model analyzes a sequence of observations of driving time series data.
Trajectory data such as velocity, acceleration, and space gap between the leader and follower drivers were used in model learning.
The results showed that by using real car following data sets, the model was able to classify normal car following driving behavior, rare events, and short-time events.
In addition, the system was able to determine the switching dynamics among different regimes by applying maximum likelihood estimates and Hamilton filter.
Moreover, the proposed model can infer regime specific characteristics such as expected duration, the probability of transferring from one regime to another, switching parameters and driving patterns.

In \cite{Ahmed2007MachineLearningToNetworkAnomalyDetection}, the authors use two networks; road traffic network and IP network.
They recorded images by six cameras over a period of four days. 
The discrete wavelet transform (DWT) algorithm is used to process the images.
The DWT is known for its ability to extract spatially localized frequency information.
The authors perform 2D DWT on every image and average the energy of transformation coefficients within each sub-band;
a sudden change in the power in the frequency content of the vector of sub-band intensities may represent an anomaly.

The IP network data constitutes a time-series of entropy of four main packet header fields (source IP address, destination IP address, source port number, and destination port number) in each of 11 $\times$ 11 backbone flows in every time step.
Anomalies represent changes in the distributions of packets.
The authors use two anomaly detection algorithms: One-Class Neighbor Machine (OCNM) and Kernel-based Online Anomaly Detection (KOAD).
OCNM uses the k$^{th}$ nearest neighbor Euclidean distance as a sparsity measure. 
KOAD is associated with a kernel function where the features corresponding to the normal traffic measurements are clustered.
The region of normality is represented using a relatively small dictionary of approximately linearly independent elements.

In \cite{Srivastava2011CoordinateMappingAnalysisVehicleTrajectory}, the authors proposed a method to detect anomalies in the trajectory of a vehicle by observing the patterns in its velocity. The authors proposed a look-up table for mapping the trajectory from image co-ordinates to a hypothetical coordinate system in which the axes are selected with respect to the road.
In this co-ordinate system, the (axial) velocity is a one dimensional quantity.
This is a spatial representation of the velocity which helps localizing the normal behavior definition.
The authors estimate the normal modes in the velocity after scaling the velocities by the average speed of the vehicle. 
The authors detect deviations from the normal velocity for individual sections of the roadway where the normal velocity was modeled using a mixture of Gaussians (obtained by training).
The authors use the shape of the trajectory to determine turns and other significant maneuvers,
an anomaly is also detected if the velocity of a vehicle falls in a path model which is inconsistent with the shape of its trajectory.
The authors characterize the shape of the trajectory using template matching in order to detect turns and other maneuvers.
Template matching was performed using sliding windows which does not require pre-segmentation of the trajectory.
Thus, it is very suitable for real-time operation.


In \cite{Johnson2011DrivingStyleRecognitionUsingSmartphoneSensorPlatform}, the authors proposed a novel system that uses dynamic time warping (DTW) and smart phone sensors (accelerometer, gyroscope, magnetometer, GPS, and video) in order to detect and record driving style activities.
The proposed system gathers related inter-axial data from multiple sensors into a single classifier.
The proposed system utilizes the Euler representation of device attitude (based on the gathered data) to aid the driving activity classification.
All the proposed processing is done entirely on the smart phone.

\section{Methodology}
\label{sec:Methodology}
In this section, we introduce the proposed model framework including model formula, parameter estimation, and characteristics. 

\subsection{Markov switching vector auto-regressive model}
MSVAR \cite{krolzig2013markov} is non-linear model which joins the vector auto-regressive models with hidden Markov chain models.
The MSVAR model builds a non-linear data model as piece-wise linear model;
this is achieved by modeling the process to be linear in each regime.
The main objective of such a model is to find the specification of each regime using variables as switches between each regime.
Such models use intercept, mean, or both used as switches.
For instance, the model presented in \cite{hamilton1989new} utilizes MSVAR with mean switches to study business cycles using the U.S. GDP series.
The major difficulty of using the mean as a switching parameter is the estimation of the switching parameters due to their interrelation and the latent variable.
On the other side, models that use the intercept as switches require less effort in estimation, it can be estimated using Monte Carlo methods.

The model introduced is based on multivariate time-series $Y = (y_{1},\dots ,y_{t})$ consisting of $t$ observations, where $y_{t}$ represents an N-dimensional vector and a stochastic process that depends on a latent discrete stochastic process, $S_t$, having discrete state-space with state variable $s_{t}$ which indicates the dominant regime at time $t$.
The reduced form of the model is presented in \cite{krolzig2013markov} and is known by MSIAH-VAR(p) as in \eqref{eq:MSVAR}.
The reduced model uses three types of switches: intercept $A_{s_{t}}^{(0)}$, regression coefficient $ A_{s_{t}}^{(i)}$, and co-variance matrix $U_{t}$.

\begin{equation} \label{eq:MSVAR}
y_{t} = A_{s_{t}}^{(0)} +  \sum \limits_{s=1}^p A_{s_{t}}^{(i)} y_{t-i} + U_{t} ,
\end{equation}
$ \hspace{10mm} \text{where} \hspace{10mm}	U_{t}|s(t) ~ \text{i.s} \ N(0,\sum \limits_{s_{t}})$.

The state variable $s_{t}$ is evolved over time as a discrete time, discrete space Markov process, assuming $s_{t} = 1, 2, \dots k$ for $k$ regimes. 
Let $M$ represents the number of appropriate regimes, so that $s_{t} \in {1,...,M}$. The conditional probability density of the observed time series vector $y_{t}$ is given by \eqref{eq:conditionalobserved} where $\theta_{M}$ is the VAR model parameter vector for regime $M$ and $Y_{t-1}$ are the observations from time $T=0$ to time $T=t-1$.

\begin{equation} \label{eq:conditionalobserved}
p(y_{t}|Y_{t-1},s_{t}) =
 \begin{cases}
  f(y_{t}|Y_{t-1},\theta_{1}) \hspace{10mm} \text{if} \ s_{t} = 1 ,\\
  \cdots \\
  f(y_{t}|Y_{t-1},\theta_{M}) \hspace{10mm} \text{if} \ s_{t} = M.
 \end{cases}
\end{equation}

The stochastic transition of states is determined by a Markov transition matrix $p$ which determines the dynamics of the switching process where $p_{i,j}$ = $Pr(s_{t} = i | s_{t-1} = j)$ is the probability of switching from state $j$ to state $i$ and $\sum \limits_{j=1}^M p_{i,j}= 1$.

The model parameter state vector is defined as follows:
\begin{equation}
\begin{split}
\theta=\{A_{1}^{(0)},A_{2}^{(0)},..A_{M}^{(0)},A_{1}^{(1)},A_{2}^{(1)}..A_{M}^{(1)}, \\
U_{1},U_{2},..,U_{M},p_{i,j}\}.
\end{split}
\end{equation}

This model parameter state vector can be estimated by maximum likelihood as in Eq. \eqref{eq:MSM_Max_likelihood}.
The estimation process is performed through the Expectation Maximization (EM) algorithm that is presented in \cite{hamilton1994autoregressive}.
EM iteratively calculates the next step of the state vector $\theta_{t+1|t}$ given the previous observation and the previous state vector using the log-likelihood function of the data.
The algorithm proceeds in two steps: expectation and maximization steps.
The expectation step uses the parameter estimated from the previous maximization step to compute both filtered probability vectors.

The likelihood is proportional to the probability of observing the data given the estimated parameter.
The minimization of the log-likelihood in Eq. \eqref{eq:MSM_Max_likelihood} can be used as an objective for parameter estimation and for comparing the different model fitting schemes.

\begin{equation} \label{eq:MSM_Max_likelihood}
\ln L = \sum \limits_{t=1}^T \ln \sum \limits_{j=1}^K
P(y_{t}|s_{t}=j,\theta).P(s_{t}=j).
\end{equation}

There are three main terminologies for gaining information about different driving regimes: driving regime inference, regime classification, and regime expected duration.
\subsubsection{Regime inference}
The objective of the regime inference process is the identification of the latent regime variable $s_{t}$ from the observations $x_{t}$. 
This process requires two main steps: filtering and smoothing.

\subsubsubsection{Filtered probability estimates}
The filter step aims at estimating $\xi_{jt}$ which represents the probability of the unobserved state vector $s_{t}$. 
The probability of being under regime $j$ at time $t$ given the model parameters is given by Eq. \eqref{eq:Model_Estimate}:
\begin{equation} \label{eq:Model_Estimate}
\xi_{jt} = P(s_{t} = j | \Omega_{t};\theta ).
\end{equation}
where $\Omega_{t}$ is the sequence of observations over time given by:
 \begin{center}
 $\Omega_{t} = \{y_{t},y_{t-1},...,y_{2},y_{1}\}$.
 \end{center}
and $\theta$ is the population parameter vector which is given by:
\begin{center}
$\theta = \{\sigma_{1},\sigma_{2},...,\sigma_{k},c_{1},c_{2}...,c_{k},\phi_{1},\phi_{2},...,\phi_{k},p_{i,j} \}$.
\end{center}
with the constraints that $\xi_{jt} > 0$ and $\sum \limits_{\forall j} \xi_{jt} = 1$.

\subsubsubsection{Smoothing}
The filter step will generate estimates for state $s_{t}$ where $t = 1....,T$ using observations up to time $t$.
The smoothing task improves regime inference by taking into consideration the future observation $t+1 < T$. 
The smoothed probability is defined as $P(s_{t}|y_{t+1,...,T})$. The smoothed algorithm is a backward filter that starts from the last observation point $t=T$.
The smoothing algorithm starts by calculating the smoothed probability of the last observation point $P(s_{t}|y_{T})$, and then iterates backward to $t=1$; the algorithm steps are as follows:

\begin{enumerate}[I]
\item The smoothed probability of the last observation point:
\begin{align*}
P(s_{t}|y_{T}) & = \sum \limits_{s_{t+1}} P(s_{t},s_{t+1}|y_{T})
\\ & = \sum \limits_{s_{t+1}} P(s_{t}|s_{t+1},y_{T}) \times P(s_{t+1}|y_{T}).
\end{align*}

\item According to the Markovian assumption, the $s_{t+1}$ depends only on $s_{t}$:
\begin{align*}
P (s_{t}|s_{t+1},y_{T}) & = P(s_{t}|s_{t+1},y_{t},y_{t+1,...,T})
\\& = \frac {P(y_{t+1,...,T}|s_{t},s_{t+1},y_{T}) \times P(s_{t}|s_{t+1},y_{t})}{P(y_{t+1,...,T}|s_{t+1},y_{t})}
\\& = P(s_{t}|s_{t+1},y_{t}).
\end{align*}

\item The calculation of the smoothed probability $P(s_{t}|y_{T})$ is done by using the last term of the previous iteration $P(s_{t+1}|y_{T})$ where

\begin{align*}
P(s_{t}|s_{t+1},y _{T}) & =  \frac{P(s_{t+1}|s_{t},y_{t}) \times P(s_{t}|y_{t})}{P(s_{t+1}|y_{t})}
\\& = \frac{P(s_{t+1}|s_{t}) \times P(s_{t}|y_{t})}{P(s_{t+1}|y_{t})}.
\end{align*}

\item The recursion is initialized with the final filtered probability vector $P(s_{t}|y_{T})$. 
The following equation shows how the future observation $y_{t+1,...,T}$ improves the inference of the unobserved state $s_{t}$.

\begin{align*}
P &(s_{t}|y_{T}) = \sum \limits_{s_{t+1}=1}^M \frac {P(s_{t+1}|s_{t}) \times P(s_{t}|y_{t})}{P(s_{t+1}|y_{T})} \times P(s_{t+1}|y_{T}).
\end{align*}

\end{enumerate}

\subsubsection{Regimes classification}
The classification of regimes begins by assigning each observation $y_{t}$ to a regime $S$.
The classification is achieved by mapping each observation to the winning regime with the highest smoothed probability as in Eq. \eqref{eq:RegimeClassification}.

\begin{equation}\label{eq:RegimeClassification}
\hat{s_{t}} =  \operatorname{arg\,max}_{1,...M} P(s_{t}=m|y_{T}).
\end{equation}

The smoothed regime probabilities are calculated using the dataset. Then each observation is assigned to the highest regime filtered probability.

\subsubsection{Regime expected duration}
The expected length of stay in a specific regime (state $i$) can be derived from the regime transition matrix; this is achieved by using the probability of staying in the same regime $P_{ii}$. 
Let $D_{i}$ be the time period in which the system stays at regime $i$.
Equation \eqref{eq:RegimeExpectedDuration} is the probability to stay time period $k$ in regime $i$.
The expected duration can be specified by the formula presented in Eq. \eqref{eq:RegimeExpectedDuration1}, and according to the formula the expected duration depends only on the transition probability $P_{ii}$ of the same regime, so the expected duration remains constant over time and the higher the transition probability, the longer the stay for the regime.

\begin{equation} \label{eq:RegimeExpectedDuration}
	P(D_{i} = k) = P_{ii}^{k-1} (1-P_{ii}).
\end{equation}

\begin{equation} \label{eq:RegimeExpectedDuration1}
	E(D_{i}) = \sum \limits_{k=0}^\infty (k \times P(D_{i} = k)) = \frac{1}{1-P_{ii}}.
\end{equation}

\subsubsection{Gipps bayesian parameter estimation}
\label{sec:GippsSampling}
The estimation of the MSVAR model is a difficult task.
Parameter estimation can be done easily for Bayesian models having known closed form posterior distributions.
The main objective of our proposed model is to adopt the MSVAR in driver behavior modeling where we cannot determine the model parameters of prior distributions directly.
MCMC can be used for finding the posterior distribution for the model parameters.
This is attributed to the ability of MCMC to generate samples from the posterior distribution.
The estimation of the proposed MSVAR model has been implemented based on Gipps sampler for posterior distribution sampling.

The Gipps sampler is a popular and efficient MCMC sampling algorithm \cite{BayesianDataAnalysis}.
The Gipps sampler is like the Metropolis Hastings (MH) component-wise implementation in sampling each dimension.
However, instead of sampling each dimension from an independent proposed distribution, Gipps samples from variable full conditional distribution $P(y_{j}|y_{-j}) = P(y_j|y_1,y_2,\dots,y_{j-1},y_{j+1},\dots,y_n)$.
The algorithm accepts all drawn samples, thus Gipps has lower computation requirements and converges faster \cite{Gipps1981behavioural}.
Like the component-wise implementation, the Gipps algorithm step-samples through each variable while the other variables are fixed.





If the target conditional distribution is belonging to standard distributions, then the sampling can directly be done from these distributions, otherwise Metropolis–Hastings algorithm can be used for sampling the target distribution. The Bayesian parameter estimation approach assumes that both of the regime $S$ and the model parameters $\theta$ are random variables. 
The Gipps sampler has been used in the inference of the state-space models, even it has been used for classifying the states without estimating the model parameters. 
The Gipps sampler can be used for sampling the parameters of the posterior distributions.


The sampler draws samples from the latent states and samples the model parameters from the full conditional distribution.
The sampler starts with sampling $\theta^{i}$ from $p(\theta|S^{i-1},\Omega_{T})$ where $\Omega_{T}$ is the observed data,
then sampling $S^{i}$ from $p(S|\theta^{i},\Omega_{T})$.
The prior specification of the state-space sampling for known number of states is followed (as presented in \cite{richardson1997bayesian}) where the standard distribution families selected for model parameters are implemented and the model parameters are conditionally independent.
The parameters prior distribution are estimated as follows:

\begin{itemize}
\item The joint transition probabilities use independent Dirichlet distribution prior $Dir(1,1,1,\dots,1)$ for each state.
\item Each regression coefficient mean $\mu_i$ has independent Gaussian prior.
\item Each regression coefficient standard deviation has gamma prior.
\end{itemize}

The presented prior specification has full state conditional distribution which follows a Dirichlet distribution as in Eq. \eqref{eq:Dir_prob} where $I \lbrace.\rbrace$ is an indicator function for indicating the current state.

\begin{equation} \label{eq:Dir_prob}
\begin{split}
p(s_{1,..,k}|\theta,\Omega) \backsim Dir(I \lbrace s_1 = 1\rbrace+1 , \hspace{2cm} \\ 
 I\lbrace s_2=1 \rbrace +1,\dots,I \lbrace s_k=1 \rbrace +1).
\end{split}
\end{equation}

The joint transition probabilities $p_{ij}$ for each state has full conditional distribution which follows a Dirichlet distribution as in Eq. \eqref{eq:Dir_tran_prob} where $n_{ij}$ is the number of transitions from state $i$ to state $j$.

\begin{equation} \label{eq:Dir_tran_prob}
p(p_{1,i,...,k}|\theta,\Omega,s) \backsim Dir(n_{11}+1,n_{12}+1,\dots,n_{1k}+1).
\end{equation}

The Gipps sampler iterates on two steps. Step (a) for updating the parameters and Step (b) for Markov chain revised as follows:

\begin{itemize}
\item Step a
\begin{itemize}
\item Update the mean by sampling from the Gaussian prior.
\item Update the standard deviation gamma prior.
\item Update the transition probabilities for each state independently by sampling from the proposed distribution in Eq. \eqref{eq:Dir_tran_prob}.
\item Update the $p(s_{1,..,k}|\theta,\Omega)$ from the proposed distribution in Eq. \eqref{eq:Dir_prob}.
\end{itemize}

\item Step b
\begin{itemize}
\item Update the filtered probability $P(s_{t} = j |\Omega,\theta)$.
\item Update the transition probabilities \\  $P(s_{t}=j|s_{t-1}=i)$.
\end{itemize}
\end{itemize}

\section{Experimentation}
\label{sec:Experimentation}


\subsection{Car following data set}
\begin{figure}[t]
\centering
\includegraphics[width = 8 cm , height = 9 cm]{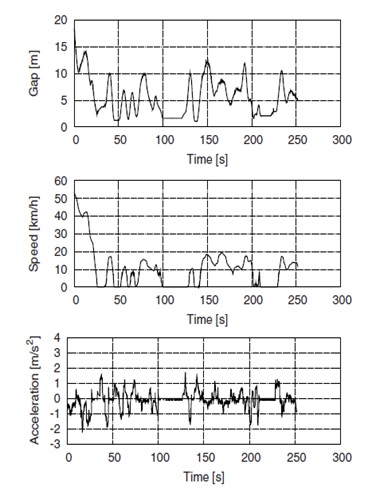}
\caption{Dataset\_1 gap distance, velocity difference, and acceleration.}
\label{Fig:dataset1}
\end{figure}

The Robert Bosch GmbH Research Group \cite{manstetten1997traffic} floating car dataset (FCD) is used to validate our model.
This dataset represents a car following behavior of vehicle speed under stop-and-go traffic conditions during an afternoon peak on a single lane in Stuttgart, Germany. 
A car with a frontal radar sensor based on a Doppler ultrasound is used to measure the relative speed and distance between a leader and a follower drivers.
The used datasets are sampled at 100 ms with duration of 250, 400, and 300 seconds.
Data set 1 gap distance, speed, and acceleration are shown in Fig. \ref{Fig:dataset1}.

These datasets have complex situations in daily urban traffic with lots of acceleration and deceleration periods.
Due to the existence of traffic lights in the recorded scenario, there are some standstill periods.
The velocity varies in the range between 0 km/h and 60 km/h.
These datasets are used in modeling, evaluating, and calibrating car following models such as the Intelligent Driver Model (IDM) in \cite{treiber2013microscopic}, the neural network models \cite{panwai2007neural}, and the state-space models \cite{hoogendoorn2007adaptive}.

\subsection{Data collection}
The driving data collection is a complex task. Most vehicle data collection experiments consist of high quality recording of driver's behavior. 
Lots of sensors have to be equipped within the vehicle equipment in order to record the various driver behavior signals. 
Sensors such as microphones, video camera, steering wheel angle, gas pedal, brake pedal, GPS, speed, acceleration, and heart rate can be used according to the objective behind the study.

The collection of driving behavior with this procedure has a high cost. 
Thus, a low cost solution using smartphones to collect car following behavior data has been introduced. 
Sensor data from both follower and leader vehicles is highly beneficial and accordingly used to fit the proposed model (using smartphones iPhone 6 and 6 plus).
We converted the GPS latitude and longitude to the actual distance between the two vehicles considering the spherical shape of the earth using the Haversine formula \cite{HaversineFormula}. 

The driving experiment has the following characteristics:
\begin{itemize}
  \item There are follower and leader drivers with predefined set of behaviors.
  \item Every driver has one smartphone in his vehicle.
  \item Every smartphone has SensorLog application running (an application for logging sensors data).
  \item All sensors are being logged; e.g., GPS, accelerometer, gyroscope, compass (location heading).
  \item Sampling rate of GPS is 1 Hz (1 sample per second).

  \item Sampling rate of acceleromoter is 100 Hz.
  \item We depend on GPS samples for localization.
  We obtain distances, velocities, and accelerations from GPS readings.
  The accelerometer does not readily provide acceleration data since it is relative to free fall.
  \item Velocity is calculated by SensorLog (using 2 consecutive GPS readings).
  \item Acceleration is calculated from the velocity difference.
  \item We need to estimate, and accordingly classify, the follower velocity based on follower acceleration, velocity difference, and gap distance.
\end{itemize}

\subsection{Adopted driving behavior model}
MSVAR models presented in Eq. \eqref{eq:MSVAR} are adopted using driving signals that represent the car following behaviors.
The signals used are presented in Eq. \eqref{eq:MSVARCarFollow} where the observation vector consists of four observation signals $y_{t} = [v_{t}, a_{t}, \Delta v_{t},h_{t}]$.
These signals represent the car following driver behavior where $v_{t}$ is the follower velocity, $a_{t}$ is the follower acceleration, $\Delta v_{t}$ is the difference in velocity between the leader and follower, and $h_{t}$ is the gap distance between the leader and follower.
The prediction of the model has $n$ interval forecasts which can be evaluated using the conditional mean $\widehat{y}_{t+n}$ and the mean square prediction error (MSPE).

The objective is to find the conditional density of $y_{t+n}$ given the model parameters and the previous observation $\Omega_{t}$. 
The prediction density given in Eq. \eqref{eq:predictionDensity} is a mixture of normals, and $p(y_{t+n}|s_{t+n} = j,\Omega_{t})$ is the probability of each predicted regime.

\begin{equation} \label{eq:MSVARCarFollow}
{
\small
\begin{split}
v_{t} =
\begin{cases}
 A_{1,1}^{(0)} + A_{1,1}^{(1)}v_{t-1} + \cdots + A_{1,1}^{(p)}v_{t-p} + U_{1,1}   \hspace{10mm}  \text{if } s_{t} = 1, \\
 ...\\
A_{1,M}^{(0)} + A_{1,M}^{(1)}v_{t-1} + \cdots + A_{1,M}^{(p)}v_{t-p} + U_{1,M}    \hspace{10mm} \text{if } s_{t} = M.
  \end{cases}
 \\
 \\
  a_{t} =
\begin{cases}
 A_{2,1}^{(0)} + A_{2,1}^{(1)}a_{t-1} + \cdots + A_{2,1}^{(p)}a_{t-p} + U_{2,1}    \hspace{10mm}  \text{if } s_{t} = 1, \\
 ...\\
 A_{2,M}^{(0)} + A_{2,M}^{(1)}a_{t-1} + \cdots + A_{2,M}^{(p)}a_{t-p} + U_{2,M}    \hspace{10mm} \text{if } s_{t} = M.
  \end{cases}
  \\
  \\
\Delta v_{t} =
\begin{cases}
 A_{3,1}^{(0)} + A_{3,1}^{(1)}\Delta v_{t-1} + \cdots + A_{3,1}^{(p)}\Delta v_{t-p} + U_{3,1}    \hspace{10mm}  \text{if } s_{t} = 1, \\
 ...\\
A_{3,M}^{(0)} + A_{3,M}^{(1)}\Delta v_{t-1} + \cdots + A_{3,M}^{(p)}\Delta v_{t-p} + U_{3,M}    \hspace{10mm} \text{if } s_{t} = M.
 \end{cases}
  \\
  \\
  h_{t} =
\begin{cases}
 A_{4,1}^{(0)} + A_{4,1}^{(1)}h_{t-1} + \cdots + A_{4,1}^{(p)}h_{t-p} + U_{4,1}    \hspace{10mm}  \text{if } s_{t} = 1, \\
 ...\\
 A_{4,M}^{(0)} + A_{4,M}^{(1)}h_{t-1} + \cdots + A_{4,M}^{(p)}h_{t-p} + U_{4,M}    \hspace{10mm}  \text{if } s_{t} = M.
  \end{cases}
\end{split}
}
\end{equation}

\begin{equation}\label{eq:predictionDensity}
\begin{split}
	p(y_{t+n}|\Omega_{t}) =  \sum \limits_{j=1}^M Pr(s_{t+n} = j)|\Omega_{t}) \\
	 \times p(y_{t+n}|s_{t+n} = j,\Omega_{t}). \\
\end{split}
\end{equation}

\begin{equation}
\begin{split}
Pr(s_{t+n} = j|\Omega_{t})	= \sum \limits_{i=1}^M Pr(s_{t+n}=j | s_{t} = i ) \\ \times Pr(s_{t} = i | \Omega_{t}).	
\end{split}
\end{equation}

\section{Results and Discussion}
\label{sec:ResultsAndDiscussion}




\subsection{Results of data collected using smartphones}
The data collected from the smartphones have been fitted using the Markov Regime Switching model presented in Eq. \eqref{eq:MSM_CarFollowing_Model}, where the driving signals selected as features are: follower velocity $v$, acceleration $a$, velocity difference $dv$, and gap distance $h$. The model is introduced in our previous work \cite{Zaky2014CarFollowingRegimeTaxonomyMarkovSwitching} with the model ability to classify the different car following driving regimes.

\begin{equation} \label{eq:MSM_CarFollowing_Model}
v_{t+1} = \phi_{1,s_{t}} a_{t}  +
\phi_{2,s_{t}} dv_{t}  +
\phi_{3,s_{t}} h_{j,t} +
\epsilon_{t}.
\end{equation}

The estimated model parameters $\sigma_{s_{k}}, \phi_{1}, \phi_{2}$ and $\phi_{3}$ are presented in Table \ref{table:RegimeSwitchingParameters} for each regime, where the log-likelihood approach is used. 
Markov transition matrix $p$ estimated, where $P(s_{t}=i|s_{t-1}=j)$ is the probability of moving from driving regime $j$ to driving regime $i$:

\begin{equation*}
p=
 \begin{pmatrix}
 	 0.94 &  0.05 &  0.01 &  0.00  \\
      0.00 &  0.85 &  0.00 &  0.01  \\
      0.00 &  0.00 &  0.81 &  0.51  \\
      0.06 &  0.10 &  0.18 &  0.48
 \end{pmatrix}
\end{equation*}

Table \ref{table:Regime_Expected_Duration} presents the observed information of each driving regime in the dataset. 
We can observe the expected duration in which a driver can stay in each regime based on Eq. \eqref{eq:RegimeExpectedDuration1}.
This means that the driver will stay for a time around the expected duration time driving in that regime.
Other characteristics of regimes are shown such as the number of occurrence which counts the samples inside each regime, the number of observations that belongs to each regime, and the percentage of driving under each regime overall time.

\begin{table}[h]
\centering
\scalebox{1}{
\begin{tabular}{l c c c c}
\hline
\multicolumn{5}{c}{\cellcolor{gray!25}Regime Characteristics} \\
\hline
\hline
State & \begin{tabular}[c]{@{}l@{}}Expected\\  duration (ms)\end{tabular}    & Occurrence& Observations &Percentage\\ \hline
Regime 1   &    17.50 &	4	& 276 	&  67\%  	\\ \hline
Regime 2   &    6.79  & 8 	& 42 	&  10\% 	\\ \hline
Regime 3   &    5.34  & 10 	& 34 	&  8.25\% 	\\ \hline
Regime 4   &    1.91  & 11 	& 60 	&  14.5\% 	\\
\hline
\hline
\end{tabular}
}
\caption{Driving regimes contained in the adopted dataset.}
\label{table:Regime_Expected_Duration}
\end{table}

\begin{table}[b]
\centering
\scalebox{1}{
\begin{tabular}{c c c c c c|}
\hline
\hline
\cellcolor{gray!25}Parameter & \cellcolor{gray!25} Regime 1 & \cellcolor{gray!25} Regime 2 &\cellcolor{gray!25} Regime 3 &\cellcolor{gray!25} Regime 4 	\\ \hline
$\sigma_{s_{k}}$ & 1.3346 & 1.2511 & 1.6234 & 1.9769  	\\ \hline
$\phi_{1}$&  1.2163 & -0.4069 & 0.0671 & -1.5901   		\\ \hline
$\phi_{2}$&  1.5228 & 1.2641 & 0.3858 & 0.9247  	\\ \hline
$\phi_{3}$& 0.6563 & 0.5074 & 0.2625 & 0.3031   		\\
\hline
\hline
\end{tabular}
}
\caption{Estimates of car following Markov regime switching model parameters.}
\label{table:RegimeSwitchingParameters}
\end{table}

We have conducted several trials and due to the noise of the sensors specially the GPS (which has up to 4 meters error with a sampling rate of 1 sample per second), our experiment is limited by driving in only three car following situations: acceleration, braking, and normal following. 
A manual tagging for the car following situations is done. The interpretation of the results is as following:
\begin{itemize}
\item Tag 0: is the acceleration behavior which the model classifies as Regime 2.
\item Tag 2: is the braking behavior which the model classifies as Regime 3.
\item Tag 4: is the stable following behavior which the model classifies as Regime 1.
\end{itemize}

\noindent The results comply with the labels taken by the follower driver that is out of 412 records (seconds):

\begin{itemize}
\item 320 records are tagged as label 4 (stable following). The model classifies 276 observations with mis-classification of 44 observations, i.e., accuracy of 86.25\%.

\item 49 records are tagged as label 0 (acceleration). The model classifies 42 observations with mis-classification of 7 observations,
 i.e., accuracy of 85.42\%.

\item 43 records are tagged as label 2 (braking). The model classifies 34 observations with mis-classification of 9 observations, 
 i.e., accuracy of 79\%.
\end{itemize}

\subsection{Results of the naturalistic driving data}
We have implemented different MSVAR models with different regimes, varying from 2 to 5 regimes and different time lag varying from 2 to 5 ms. 
The maximum log-likelihood values are shown in Table \ref{table:VARLogLikelihood} for different lags and regimes. 
As shown in Table \ref{table:VARLogLikelihood} the best model fit is for lag 1 having 5 regimes with the maximum likelihood value. 
The using of the lag can be useful for modeling the driver sensitivity factor which is presented in all GM models \cite{rakha2009simplified}.
The following driver responds to the leading driver action by acceleration or deceleration depending on the driver perception, reaction time, and driver sensitivity factors.
An average reaction time is estimated in a range between 1.0 to 2.2 seconds, and an average driver sensitivity factor of 0.37 second as introduced in \cite{may1990traffic}.

\renewcommand{\arraystretch}{1.5}
\begin{table}[b]
\centering
\scalebox{0.8}
{
\begin{tabular}{| c | c | c | c | c | c |}
\hline
\hline
\multicolumn{6}{ | c | }{\cellcolor{gray!25}Dataset\_1} \\
\hline
\backslashbox{Regime}{Lag}   & 1 & 2 & 3 & 4 & 5 \\ \hline
2 &  10036.68 &  10642.67 & 10739.58 & \cellcolor{gray!25} \bfseries 11057.45  & \cellcolor{gray!25} \bfseries 11090.25  \\ \hline
3 &  9908.593 & \cellcolor{gray!25} \bfseries 10869.07  & \cellcolor{gray!25} \bfseries 10992 	 & 10837.91  &  10889.22 \\ \hline
4 &  10151.84 & 10400.08  &	10981.64 & 10515.56	 & 10499.64  \\ \hline
5 &  \cellcolor{gray!25} \bfseries 10377.99 &	10244	  & 10359.58 & 10341.47  &	10005.32 \\ \hline
6 &  10191.81 &	9348.61	  & 9985.96	 & 10054.52	 & 10119.44  \\
\hline
\hline
\end{tabular}
}
\caption{MSVAR log likelihood values.}
\label{table:VARLogLikelihood}
\end{table}

Most VAR models are estimated using symmetric lags, i.e., the same lag length is used for all variables of the model. 
The model lag length $P = 1,..., p$ can be determined based on a specific selection criteria. 
Model selection criteria such as Akaike Information Criterion (AIC), Bayesian Information Criterion (BIC), and Hannan-Quinn Information Criterion (HQC) are used.
The selection of an inappropriate lag length may affect the model performance and fitting. 

\begin{figure}[!h]
    \centering
    \begin{subfigure}[b]{0.5\textwidth}
        \includegraphics[scale=0.5]{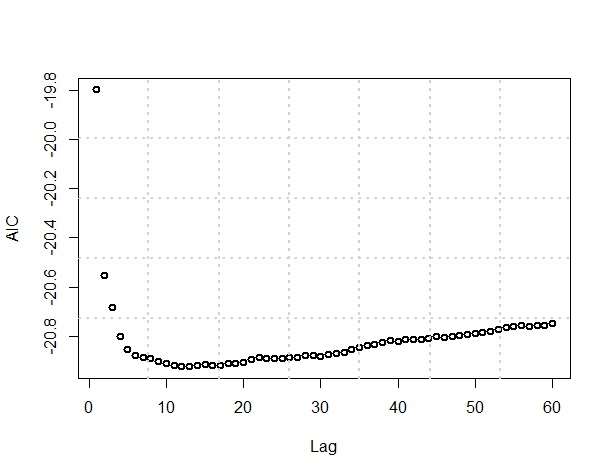}
        \caption{AIC values for lag between 1 and 60 ms.}
    \end{subfigure}
    \begin{subfigure}[b]{0.5\textwidth}
        \includegraphics[scale=0.5]{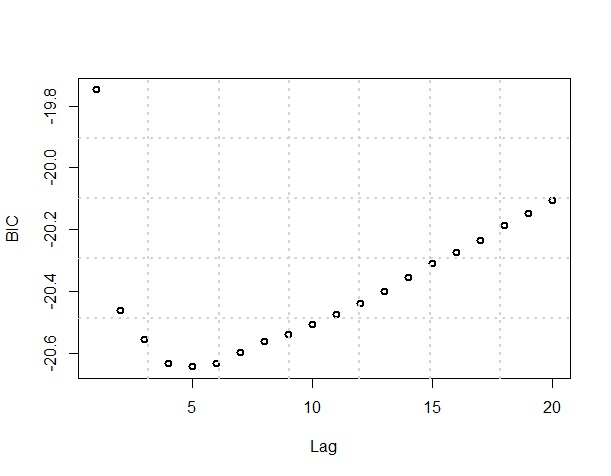}
        \caption{BIC values for lag between 1 and 20 ms.}
    \end{subfigure}
    \begin{subfigure}[b]{0.5\textwidth}
        \includegraphics[scale=0.5]{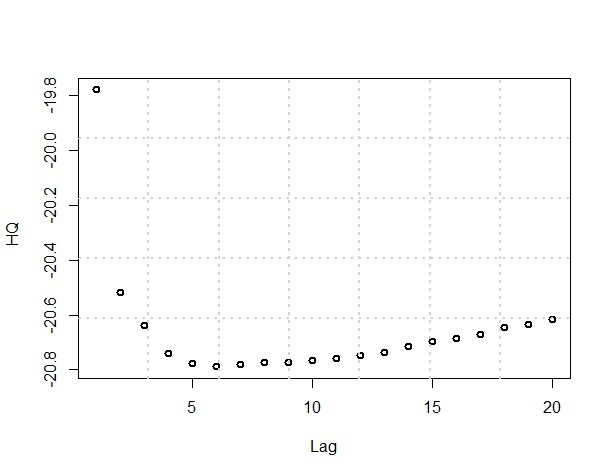}
        \caption{HQ values for lag between 1 and 20 ms.}
    \end{subfigure}
    \caption{MSVAR(P) model selection criteria.}
\label{fig:SelectionCriteria}
\end{figure}

The results of applying each criteria on the adopted dataset are shown in Fig. \ref{fig:SelectionCriteria}. 
This figure shows the results for different lags between 1 and 60 ms.
This helps us to exploit the history of the driving situation in the model. 
The dataset is sampled at 10 Hz, thus 1 lag means 0.1 second while 60 lag means 6 seconds. 
The minimum values of the three selection criteria falling between 5 and 13 (0.5 and 1.3 seconds) can be observed which represent the different reaction times of the driver.

The MSVAR model presented in Table \ref{table:VARLogLikelihood} is used for fitting the naturalistic driving data. 
The presented results are for the dataset which contains 2529 observations. 
As shown in the table, the best fitting is for the model with a lag of 5 and 2 regimes. 
The lag selection criteria described above uses 2520 observations, and the remaining 9 observations have been used for models forecasting evaluation.

The forecasting results based on Eq. \eqref{eq:predictionDensity} represent a comparison of the prediction performance of the two selected models. 
The highest max log-likelihood model is selected. 
The first model has 1 lag and 5 regimes Model I (p=1,r=5), while the second model has 5 lags and 2 regimes Model II (p=5,r=2). 
Table \ref{table:ModlesMSE} shows a comparison between the Mean Square Error (MSE) of the observation vector for the two models. 
The error is calculated between the dataset observed values and the predicted values for 9 samples (0.9 second). 
As shown, Model II has a lower MSE for all observation vector elements.

\begin{table}[!h]
\centering
\scalebox{0.8}
{
\begin{tabular}{| c | c | c | c | c |}
\hline
\hline
\cellcolor{gray!25}\backslashbox{Model}{MSE}   & $a$ & $\Delta v$ & $h$ & $v$  \\ \hline
I (p=1,R=5) 	& 0.1788184	&	0.013395644	&	0.05679842	&	0.02309714 \\ \hline
II (p=5,R=2) 	& 0.1481768	&	0.010264159	&	0.05339007	&	0.01554496 \\
\hline
\hline
\end{tabular}
}
\caption{Mean square error of the two MSVAR models.}
\label{table:ModlesMSE}
\end{table}

\begin{figure}[!h]
    \centering
    \begin{subfigure}[b]{0.5\textwidth}
        \includegraphics[scale=0.5]{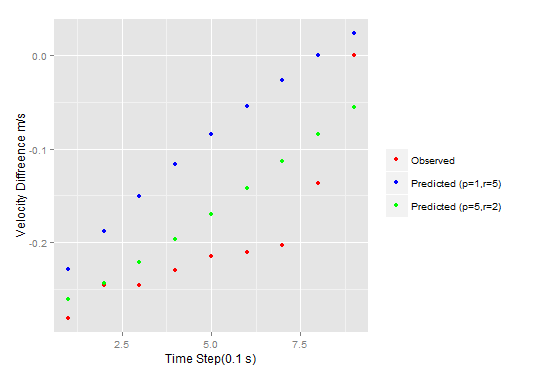} 
        \caption{DVP}
    \end{subfigure}
    \begin{subfigure}[b]{0.5\textwidth}
        \includegraphics[scale=0.5]{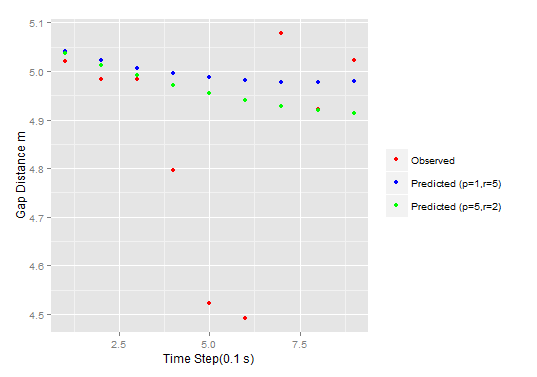}  
        \caption{HP}
    \end{subfigure}
    \begin{subfigure}[b]{0.5\textwidth}
        \includegraphics[scale=0.5]{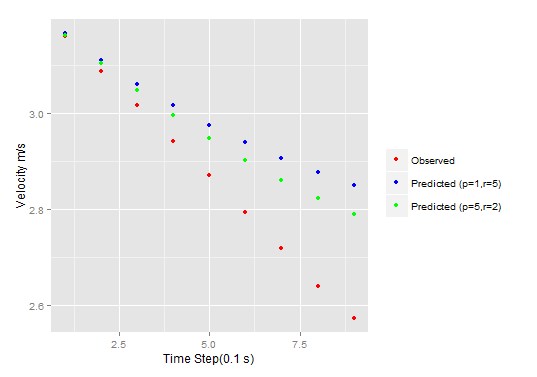}  
        \caption{VP}
    \end{subfigure}
    \caption{MSVAR model predicted values of car following observations.}
\label{fig:PredectedResults1}
\end{figure}

Fig. \ref{fig:PredectedResults1} presents the prediction of the two models for each point. 
The red points represent the naturalistic real driving data samples, blue points represent Model I, and green points represent Model II. 
Both models are accurate for the first forecasting steps, afterwards the models start deviating due to the accumulated error of the forecasting process (as shown in the velocity and velocity difference figures). 
The models are able to predict not only the driver behavior represented by its observation (velocity and acceleration), but also the entire driving situation represented by the relation with the leading vehicle observation (gap distance and velocity difference).

\subsection{MSVAR versus PrARX}
The proposed MSVAR framework, as a switching linear regression model for driving behavior, has features listed in Table \ref{table:PrARXvsMRS}. 
The table also lists the features of PrARX as a switching framework based on linear regression. 
Both frameworks are able to provide behavior/mode segmentation extracted from driving signals. 
PrARX uses K-means clustering and the proposed framework uses a probabilistic classifier based on selecting the maximum state filtering value for each observation. 
The probabilistic classification approach is able to identify the membership probability of a new observation to each state (finds the best state that represents the observation with the highest probability). 
The advantage of probabilistic classifier over non-probabilistic one is that the former behaves like a confidence weighted classifier which helps avoiding error propagation. 
The classifier adds value for the driving behavior problem by allowing a smooth transition between each regime allowing a mixed mode representation to understand the current behaviors that the driver may behave. 

\begin{table}[!h]
\centering
\begin{tabular}{|p{\dimexpr 0.18\linewidth-2\tabcolsep}|p{\dimexpr 0.4\linewidth-2\tabcolsep}|p{\dimexpr 0.42\linewidth-2\tabcolsep}|}
\hline
\hline
 		 &  PrARX & Proposed MSVAR framework 	\\ 
\hline
Behavior (Mode) segmentation &  Extension to k-means clustering & Probabilistic classification \\ 
\hline
Learning process & Cannot simultaneously classify and estimate &  Classify and estimate in a single recursive process \\ 
\hline
Parameter estimation method & Steepest descent & ML (Hamilton) and Bayesian (Gipps Sampling) \\ 
\hline
Computational cost	 & High due to the processing of the classification and estimation independently &  MCMC minimize computational time \\ 
\hline
Assumptions & Human driving behavior does not exhibit an abrupt change & Abnormal behavior detected and can capture short time events - behavior and switching processes are stationary \\ 
\hline
Mixed mode & Can & Can \\ 
\hline
\hline
\end{tabular}
\caption{Main features of PrARX model and the proposed MSVAR framework}
\label{table:PrARXvsMRS}
\end{table}

The learning process of PrARX has two stages; behavior classification and parameters estimation. 
PrARX framework cannot simultaneously classify and estimate. 
The proposed framework classifies and estimates in a recursive process by using Hamilton filter for classification and maximum likelihood for parameter estimation. 
Computational cost of PrARX is higher due to the processing of classification and estimation independently. 
For the proposed model, a Bayesian parameter estimation based on MCMC minimizes computational time over Expectation Maximization (EM). 
As a future work, performance measurements for the framework evaluation will be conducted. 

PrARX has a major assumption for relaxing the problem of parameter estimation. 
The assumption is that human driving behavior does not exhibit an abrupt change which allows the framework to use the parameters of the previous behavior as initial values for the next one. 
The proposed MSVAR framework does not have this assumption as the model parameter estimation process generates independent parameters for each regime, which allows the framework to handle abnormal behavior. 
The proposed MSVAR framework has an assumption that the driver behavior is stationary, so the stochastic process representing each regime and the switching process (transition matrix) have the same parameters over time. 
Releasing this assumption requires a variable time transition probabilities and more complex stochastic models for driver behavior modeling such as Gaussian Mixture Models (GMM).
      
\section{Conclusions and Future Work}
\label{sec:ConclusionsAndFutureWork}

The prediction of driving behavior based on multiple Markov Switching Variable Auto-Regression (MSVAR) is introduced. 
More than one model are implemented with different parameters (lag and regime) and with different evaluation criteria (AIC, BIC, HQC). 
The best fitted models are selected for the prediction process. 
Additionally, the model is capable of fitting driving data and data segmentation into regimes by estimating the different driving behavior change points. 
One limitation is the long calibration time of the model parameters. 
This is attributed to the learning of each model (depending on the model configuration the learning takes up to 3 days). 
The best fitted model is achieved at lag of 13 (1.3 s). 
We have implemented models with lags down to 5 (0.5 seconds). 
The computational efficiency of the prediction is reasonable; it takes only few seconds, however, it needs more adaptation to be more accurate. 
We also presented a low-cost data collection solution using smartphones validated with another naturalistic driving data set for predicting the driver behavior for short periods of time. 
The proposed driver behavior detection model can potentially be used in systems such as accident prediction and driver safety.

\section{Acknowledgements}
This work is mainly supported by the Ministry of Higher Education (MoHE) of Egypt through PhD fellowship awarded to Dr. Ahmed Zaky.
This work is supported in part by the Science and Technology Development Fund (STDF), Project ID 12602 ''Integrated Traffic Management and Control System'', and by E-JUST Research Fellowship awarded to Dr. Mohamed A. Khamis.

\bibliographystyle{vancouver-authoryear.bst}
\bibliography{references}

\end{document}